# Benchmarking Deep Learning Architectures for Predicting Readmission to the ICU and Describing Patients-at-Risk

Sebastiano Barbieri*[1], James Kemp[1], Oscar Perez-Concha[1], Sradha Kotwal[2,5], Martin Gallagher[2], Angus Ritchie[3,4], Louisa Jorm[1]

[1] Centre for Big Data Research in Health, University of New South Wales, NSW, Australia

[2] The George Institute for Global Health, University of New South Wales, NSW, Australia

[3] Concord Clinical School, The University of Sydney, Sydney, NSW, Australia

[4] Health Informatics Unit, Sydney Local Health District, Camperdown, NSW, Australia

[5] Prince of Wales Hospital, Randwick, NSW, Australia

* Corresponding author: Sebastiano Barbieri

Centre for Big Data Research in Health, University of New South Wales, Level 2 AGSM Building (G27), UNSW 2052 NSW, Australia.

E-mail: s.barbieri@unsw.edu.au.

ABSTRACT

**Objective:** To compare different deep learning architectures for predicting the risk of readmission within 30 days of discharge from the intensive care unit (ICU). The interpretability of attention-based models is leveraged to describe patients-at-risk.

**Methods**: Several deep learning architectures making use of attention mechanisms, recurrent layers, neural ordinary differential equations (ODEs), and medical concept embeddings with time-aware attention were trained using publicly available electronic medical record data (MIMIC-III) associated with 45,298 ICU stays for 33,150 patients. Bayesian inference was used to compute the posterior over weights of an attention-based model. Odds ratios associated with an increased risk of readmission were computed for static variables. Diagnoses, procedures, medications, and vital signs were ranked according to the associated risk of readmission.

**Results**: A recurrent neural network, with time dynamics of code embeddings computed by neural ODEs, achieved the highest average precision of 0.331 (AUROC: 0.739, $F_1$-Score: 0.372). Predictive accuracy was comparable across neural network architectures. Groups of patients at risk included those suffering from infectious complications, with chronic or progressive conditions, and for whom standard medical care was not suitable.

**Conclusions**: Attention-based networks may be preferable to recurrent networks if an interpretable model is required, at only marginal cost in predictive accuracy.

# Introduction

Modern machine learning algorithms such as artificial neural networks with multiple hidden ("deep") layers can extract relevant features from medical data and make predictions for previously unseen patients. Examples of successful applications of deep learning techniques in the medical domain include the classification of skin cancer images with accuracy comparable to dermatologists [1], the prediction of cardiovascular risk factors from retinal fundus photographs [2], and the prediction of diseases including severe diabetes, schizophrenia, and various cancers based on information contained in electronic medical records (EMR) [3]. Nonetheless, integration of machine learning assistants into clinical processes remains scarce and actual benefits for patient care are yet to be demonstrated [4].

Attention layers in deep neural networks enable them to focus on a subset of inputs (features) and respond accordingly, increasing model performance. They also offer much-needed model interpretability, by identifying which components of the input are attended to at any point in time. Current deep learning architectures used for risk prediction based on EMR data generally employ attention layers on top of recurrent layers [5-7] (e.g. Long Short-Term Memory, LSTM [8], or Gated Recurrent Units, GRU [9]). While these models yield state of the art results in prediction accuracy, the use of recurrent layers is associated with several drawbacks: interpretation of results is hampered by outputs being a nonlinear combination of current input and current memory state; lack of set-invariance (i.e. outputs differ based on the specific sequence of timestamped variables within the EMR, even if these variables were recorded at the same time) [10]; and long training times due to difficulties in parallelizing these sequential algorithms. Neural networks relying entirely on attention mechanisms have been proposed as an alternative to overcome the limitations of recurrent neural networks, with comparable or improved accuracy on several language processing tasks [11] and when used for risk predictions based on EMR data [12].

In their basic form, neither recurrent layers nor attention mechanisms are tailored to process time-series sampled at irregular time intervals, such as the diagnosis and procedure codes contained in EMRs. A wide variety of approaches to address this issue have been proposed in recent literature, ranging from simply adding or appending time-related information to the numerical vectors ("embeddings") used to represent timestamped codes [5], to modifying the internal workings of recurrent cells using exponential time-decay functions [13-15] or ordinary differential equations (ODEs) [16-18]. Similarly, code embeddings may be determined end-to-end while optimising the other parameters of the network, or in an unsupervised pre-processing step that considers time-related information, e.g. through medical concept embeddings (MCEs) with time-aware attention [19]. An additional option may be the use of neural ODEs to describe how the embedding (intuitively, the "meaning") of a medical code changes over time: codes related to a chronic condition, such as diabetes, will often maintain their relevance over years, whereas others may quickly become unimportant for prediction purposes. Amidst all these options, comprehensive work comparing neural network architectures for risk prediction based on EMR data is currently limited.

Predicting a patient's risk of readmission within 30 days of discharge from the intensive care unit (ICU) presents an exemplar application for machine learning models using longitudinal EMR data. Readmission to the ICU is an adverse outcome experienced by approximately 10% of critically ill patients following discharge [20,21] and may be an indicator of poor or incomplete medical care [21-24]. If patients at high risk of readmission can be identified, appropriate interventions, such as careful patient evaluation before discharge, planning for proper placement of the patient after discharge, and a safe and thorough handover of patient care between healthcare providers, can be implemented [25]. Readmission to the ICU also represents a major source of avoidable costs for the health care system, as up to 30% of total hospital costs and 1% of the US gross national product are directly linked to ICU expenses [26]. Points-based scores such as the Acute Physiology and Chronic Health (APACHE) score [27], the Simplified Acute Physiology Score (SAPS) [28], and the Oxford Acute Severity of Illness Score (OASIS) [29] are routinely used in ICUs to evaluate severity of illness and predict mortality risk; they may also be useful in predicting the risk of readmission [30]. However, a recent study comparing several scores used to predict the risk of readmission within 48 hours from discharge determined only moderate discrimination power (area under the receiver operating characteristic curve, AUROC, between 0.65 and 0.67) [31]. It is plausible that the application of novel machine learning algorithms to EMR data could lead to more accurate predictions.

Using the example of predicting the risk of readmission within 30 days of discharge from the ICU, the aims of the present study are: 1) to evaluate the feasibility of using neural ODEs to model how the predictive relevance of recorded medical codes changes over time; 2) to perform a comprehensive comparison of deep learning models that have been proposed for processing time-series sampled at irregular intervals, including MCEs, neural ODEs, attention mechanisms, and recurrent layers; and 3) to leverage the interpretability of attention layers combined with Bayesian deep learning to gain a better understanding of intensive-care patients at increased risk of readmission.

## Methods

### Study Population

The algorithms were evaluated on the publicly available MIMIC-III data set (ethics approval was not required) [32]. This data set comprises deidentified health data associated with 61,532 ICU stays and 46,476 critical care patients at Beth Israel Deaconess Medical Center in Boston, Massachusetts between 2001 and 2012.

The supervised learning task consists of predicting, for a given ICU stay, whether the patient will be readmitted to the ICU within 30 days from discharge. Patients were excluded if they died during the ICU stay (N=4,787 ICU stays), were not adults (18 years old or older) at the time of discharge (N=8,129 ICU stays) or died within 30 days from discharge without being readmitted to the ICU (N=3,318 ICU stays). The final data set comprised 45,298 ICU stays for 33,150 patients, labelled as either positive (N=5,495) or negative (N=39,803) depending on whether a patient did or did not experience readmission within 30 days from discharge. To develop and evaluate the algorithms, patients were subdivided randomly into training and validation (90%) and test sets (10%). This subdivision was based on patient identifiers and not on ICU stay identifiers to prevent information leaks between data sets (since the prediction is based on the entire clinical history of a patient).

### Model Variables

The EMR of a patient can be represented as a set of static variables and timestamped codes. In the present study, static variables included the patient's gender, age, ethnicity, insurance type, marital status, the previous location of the patient prior to arriving at the hospital (admission location), and whether the patient was admitted for elective surgery. Both length of ICU stay and length of hospital stay prior to ICU admission were recorded. An additional static variable was given by the number of ICU admissions in the year preceding the considered index ICU stay.

Data types of timestamped codes included international classification of diseases and related health problems (ICD-9) diagnosis and procedure codes, prescribed medications, and patient vital signs. All diagnosis and procedure codes in the clinical history of a patient were considered for predictive purposes; however, prescribed medications and recorded vital signs were restricted to the ICU stay of interest. Following the OASIS severity of illness score [29], assessed vital signs comprised the Glasgow Coma Scale score (sum of eye response, verbal response, motor response components), heart rate, mean arterial pressure, respiratory rate, body temperature, urine output, and whether the patient necessitated ventilation. Continuous measurements of vital signs were categorised in the same manner as in OASIS and assigned corresponding codes [29]. To reduce redundant information, whenever the same vital sign-related code was recorded consecutively more than once, only the latest observation was kept in the data.

Elapsed times, measured in days, associated with diagnosis and procedure codes were based on the date and time of discharge from the corresponding hospital admission. Elapsed times, measured in hours, associated with medications and vital signs were based on the date and time of prescription start and measurement, respectively. In the present study, the simplifying assumption is made that diagnosis and procedure codes are available immediately at the time of discharge from the ICU. Categorical values of static variables or timestamped codes associated with less than 100 ICU stays were re-labelled as "other".

## Artificial Neural Network Architectures

Several "deep" neural network architectures for predicting a patient's risk of readmission to the ICU were implemented and compared. To make this comparison as fair as possible, all architectures shared a similar high level structure: 1) timestamped codes were mapped to vector embeddings; 2) numerical scores associated with diagnosis and procedure codes, and with medication and vital sign codes, were computed using attention mechanisms and/or recurrent layers; 3) these scores were concatenated with the static variables and passed on to a "logistic regression layer" (i.e. a fully connected layer with a sigmoid activation function). Further details about individual network components are reported in the following sections.

### *Embeddings*

Diagnosis and procedure codes, as well as medication and vital sign codes, were mapped to corresponding "embeddings" (real-valued vectors). The size of these embeddings was set proportional to the fourth root of the total number of codes in the dictionary (diagnoses/procedures and medications/vital signs were processed separately since they were measured on different time scales) [7]. Time-aware code embeddings were computed in three different manners. A first approach used MCEs with time-aware attention [19]. MCEs are based on the continuous bag-of-words model [33], but instead of using fixed-sized temporal windows to determine a code's context, attention mechanisms learn the temporal scope of a code together with its embedding. A second approach optimised an embedding matrix at the same time as the other parameters of the network and, optionally, concatenated the elapsed times to the resulting vectors. A third approach optimised an embedding matrix at the same time as the other parameters of the network and modelled the dynamics in time of the computed embeddings using neural ODEs [16-18]. More in detail, the embedding of a code at time zero (i.e. at the time of discharge from the ICU) was stored in the embedding matrix whereas the embedding of a code recorded before discharge was computed by solving an initial value problem where derivatives with respect to time were approximated by a multilayer perceptron.

*Attention and/or Recurrent Layers*

The sequence of code embeddings associated with a patient is usually of arbitrary length and needs to be integrated into a fixed-size vector for further processing. Attention mechanisms, such as dot-product attention [34], compute a weighted average of the code embeddings, where a higher weight is assigned to the most relevant codes. Alternatively, recurrent layers iteratively process an input sequence of codes and, at each iteration, update an internal memory state and generate an output vector [8,9]. Information may be integrated for further processing by using either the final memory state of the recurrent cell or by applying an attention mechanism to the set of output vectors [5-7]. Specifically, in this work, recurrent cells were implemented using bi-directional gated recurrent units [9]. Time-related information was taken into account by concatenating the time differences between observations to the embedding vectors, by applying an exponential decay proportional to the time differences between observations to the internal memory state of the recurrent cell [13-15] or by modelling the dynamics in time of the internal memory state using neural ODEs [16-18]. To aid subsequent interpretation without altering network capacity, the fixed-size vectors produced by attention mechanisms and/or recurrent layers were reduced to two scalar-valued scores (one related to diagnoses/procedures and one related to medications/vital signs) using fully connected layers with a linear activation function.

*Logistic Regression Layer*

The computed diagnoses/procedures and medications/vital signs scores were concatenated to the vector of static variables and passed to a fully connected layer with a sigmoid activation function. The output of the network corresponds to the risk of readmission to the ICU within 30 days from discharge.

*Architectures*

The following neural network architectures were compared for predicting readmission to the ICU:

- **ODE + RNN + Attention:** dynamics in time of embeddings are modelled using neural ODEs, embeddings are passed to RNN layers, dot-product attention is applied to RNN outputs.
- **ODE + RNN:** dynamics in time of embeddings are modelled using neural ODEs, embeddings are passed to RNN layers, the final memory states are used for further processing.
- **RNN (ODE time decay) + Attention:** embeddings are passed to RNN layers with dynamics in time of the internal memory states modelled using neural ODEs, dot-product attention is applied to RNN outputs.
- **RNN (ODE time decay):** embeddings are passed to RNN layers with dynamics in time of the internal memory states modelled using neural ODEs, the final memory states are used for further processing.
- **RNN (exp time decay) + Attention:** embeddings are passed to RNN layers with internal memory states decaying exponentially over time, dot-product attention is applied to RNN outputs.
- **RNN (exp time decay):** embeddings are passed to RNN layers with internal memory states decaying exponentially over time, the final memory states are used for further processing.
- **RNN (concatenated Δtime) + Attention:** embeddings are concatenated with time differences between observations and passed to RNN layers, dot-product attention is applied to RNN outputs.

- **RNN (concatenated Δtime):** embeddings are concatenated with time differences between observations and passed to RNN layers, the final memory states are used for further processing.
- **ODE + Attention:** dynamics in time of embeddings are modelled using neural ODEs, dot-product attention is applied to the embeddings.
- **Attention (concatenated time):** embeddings are concatenated with elapsed times, dot-product attention is applied to the embeddings.
- **MCE + RNN + Attention:** MCE is used to compute the embeddings, embeddings are passed to RNN layers, dot-product attention is applied to RNN outputs.
- **MCE + RNN:** MCE is used to compute the embeddings, embeddings are passed to RNN layers, the final memory states are used for further processing.
- **MCE + Attention:** MCE is used to compute the embeddings, dot-product attention is applied to the embeddings.

The dimension of the internal memory state of RNN cells was set equal to the dimension of the input embeddings. Similarly, the dimension of the hidden representation of embeddings when computing dot-product attention was left unchanged. Derivatives with respect to time used to implement neural ODEs were approximated by a multilayer perceptron with three hidden layers of constant width equal to the size of the input. The Euler method was used as ODE solver.

An overview of the considered neural network architectures is presented in Figure 1. For completeness, the deep learning approaches were also compared with a **logistic regression** model using all static variables and the most recent vital signs for each patient as covariates.

Interpretation of Attention-based Models

For the proposed neural network architectures, the weights of the final fully connected layer can be used to determine the impact of static variables and timestamped codes on estimated risk. As in traditional logistic regression, these weights can be interpreted as increases in log-odds for unplanned early ICU readmission if the corresponding static variables or scores are increased by one unit.

It is also of interest to determine which codes (i.e. diagnoses, procedures, medications, vital signs) are associated with a prediction of high risk. Dot-product attention computes a weighted average of embedded codes; fully connected layers are then used to output scores associated with diagnoses/procedures and medications/vital signs. By passing single codes (i.e. the rows of the embedding matrix) to the fully connected layers computing these scores, it is possible to associate each code with a score. The higher the score, the higher the risk of ICU readmission when a patient's EMR contains that code.

To estimate Bayesian credible intervals around network weights and computed risk scores, the posterior distribution of weights was approximated using stochastic variational inference with mean-field approximation [35,36]. In the present study, the variational posterior is assumed to be a diagonal Gaussian distribution and is estimated using the Bayes by Backprop algorithm [37]. Following the original paper, a priori sparsity of the network weights is encouraged by formulating the prior distribution as a scale mixture of two zero-mean Gaussian densities with standard deviations of $\sigma_1 = 1$ and $\sigma_2 = e^{-6}$, respectively, and mixture weight $\pi = 0.5$ [37]. After the posterior distribution has been computed, 95% credible intervals around network weights (or combinations thereof) can be estimated by repeated sampling. Sampling of network weights may also be used to compute credible intervals around the risk prediction for a given patient.

## Training

To compare the classification accuracy of the considered neural network architectures, maximum likelihood estimates of network parameters were obtained using a log-loss cost function on the training data, extensive use of dropout with 50% probability after each embedding, RNN, and attention layers [38], and stochastic gradient descent with an Adam optimizer (batch size of 128 and learning rate of 0.001) [39]. Class imbalance was taken into consideration by assigning a proportionally higher cost of misclassification to the minority class [40]. Training was terminated after 80 epochs since overfitting of the training data started to become apparent with additional training epochs (based on average precision on the validation data). For interpretation purposes, Bayes by Backprop was used to train the "Attention (concatenated time)" neural network architecture on the entire data set, terminating if the loss function (the expected lower bound) did not decrease for 10 consecutive epochs.

## Statistical Analysis

Baseline characteristics were determined for the analysed patient population. The prediction accuracy of each considered algorithm was evaluated based on average precision, AUROC, $F_1$-Score, sensitivity, and specificity. Average precision may reflect algorithmic performance on imbalanced data sets better than AUROC as it does not reward true negatives [41,42]. The $F_1$-Score was maximised over different threshold values on risk predictions. Sensitivity and specificity were computed by maximising Youden's J statistic [43]. 95% confidence intervals associated with each metric were computed by bootstrapping, i.e. by sampling the test set with replacement 100 times and re-evaluating the models each time [44]. Since the bootstrap estimator assumes the resampling of independent events, sampling was based on patient identifiers rather than on ICU stay identifiers.

Training the "Attention (concatenated time)" network using Bayes by Backprop allowed computation of odds ratios (OR) associated with static variables and ranking of the timestamped codes (diagnoses, procedures, medications, and vital signs) according to their associated average scores (a high positive score corresponds to increased risk of readmission to the ICU); corresponding 95% credible intervals were determined using 10,000 network samples.

Software was implemented in Python using Scikit-learn [45] and PyTorch [46]; the developed algorithms are publicly available at https://github.com/sebbarb/time_aware_attention.

## Results

### Baseline Characteristics

Baseline characteristics of the analysed patient population are reported in Supplementary Table S1. In total, the models were trained using 23 static variables, 992 unique ICD-9 diagnosis codes, 298 unique ICD-9 procedure codes, 586 unique medications, and 32 codes related to vital signs. The embedding dimension was 12 for diagnoses and procedures and 10 for medications and vital signs. Each patient's EMR contained at most 552 ICD-9 diagnosis and procedure codes and 392 medications and vital sign codes associated with the current ICU stay.

## Model Comparison

Average precision, AUROC, F$_1$-score, sensitivity, and specificity for the considered deep learning architectures and the logistic regression model are reported in Table 1. The highest average precision of 0.331 was obtained by the ODE + RNN model. In general, the predictive accuracy of neural networks was considerably higher than for the logistic regression baseline model (average precision of 0.257). Models with a recurrent component (average precision range: 0.298-0.331) performed slightly better than models based solely on attention layers (average precision range: 0.269-0.294). Applying an attention layer to the outputs of RNNs at each time step instead of directly using the final memory state of the RNN, and similarly learning code embeddings end-to-end instead of using pre-trained MCE, increased network capacity but only lead to occasional, and marginal, improvements in predictive accuracy.

Table 1. Summary statistics (mean, [95% confidence interval]) for the different algorithms used to predict readmission within 30 days of discharge from the intensive care unit. ODE: Ordinary Differential Equation; RNN: recurrent neural network; MCE: medical concept embedding; AUROC: area under the receiver operating characteristic.

|  | Average Precision | AUROC | F$_1$-Score | Sensitivity | Specificity |
|---|---|---|---|---|---|
| ODE + RNN + Attention | 0.314 [0.306,0.321] | 0.739 [0.736,0.741] | 0.376 [0.371,0.381] | 0.685 [0.666,0.704] | 0.677 [0.658,0.696] |
| ODE + RNN | 0.331 [0.323,0.339] | 0.739 [0.737,0.742] | 0.372 [0.367,0.377] | 0.672 [0.659,0.686] | 0.697 [0.683,0.711] |
| RNN (ODE time decay) + Attention | 0.316 [0.307,0.324] | 0.743 [0.741,0.746] | 0.375 [0.370,0.379] | 0.648 [0.641,0.656] | 0.733 [0.726,0.739] |
| RNN (ODE time decay) | 0.300 [0.293,0.308] | 0.741 [0.738,0.744] | 0.372 [0.367,0.376] | 0.710 [0.698,0.722] | 0.667 [0.655,0.679] |
| RNN (exp time decay) + Attention | 0.320 [0.312,0.328] | 0.748 [0.745,0.751] | 0.377 [0.372,0.382] | 0.704 [0.692,0.715] | 0.680 [0.668,0.692] |
| RNN (exp time decay) | 0.304 [0.297,0.311] | 0.735 [0.732,0.738] | 0.368 [0.363,0.373] | 0.707 [0.700,0.714] | 0.670 [0.663,0.676] |
| RNN (concatenated ∆time) + Attention | 0.312 [0.303,0.320] | 0.741 [0.739,0.744] | 0.368 [0.363,0.372] | 0.687 [0.680,0.695] | 0.688 [0.681,0.696] |
| RNN (concatenated ∆time) | 0.311 [0.303,0.320] | 0.739 [0.737,0.742] | 0.364 [0.359,0.369] | 0.698 [0.692,0.704] | 0.688 [0.684,0.693] |
| ODE + Attention | 0.294 [0.285,0.302] | 0.717 [0.714,0.720] | 0.333 [0.328,0.339] | 0.776 [0.768,0.784] | 0.554 [0.548,0.560] |
| Attention (concatenated time) | 0.286 [0.277,0.295] | 0.711 [0.709,0.714] | 0.330 [0.325,0.334] | 0.700 [0.686,0.714] | 0.614 [0.601,0.628] |
| MCE + RNN + Attention | 0.317 [0.308,0.325] | 0.736 [0.734,0.739] | 0.373 [0.369,0.378] | 0.630 [0.622,0.638] | 0.744 [0.738,0.749] |
| MCE + RNN | 0.298 [0.291,0.306] | 0.727 [0.724,0.730] | 0.361 [0.357,0.366] | 0.654 [0.645,0.663] | 0.706 [0.697,0.715] |
| MCE + Attention | 0.269 [0.261,0.278] | 0.689 [0.686,0.692] | 0.312 [0.308,0.316] | 0.686 [0.676,0.695] | 0.616 [0.607,0.625] |
| Logistic Regression | 0.257 [0.248,0.266] | 0.659 [0.656,0.663] | 0.296 [0.291,0.300] | 0.606 [0.597,0.615] | 0.647 [0.639,0.655] |

Interpretation of Attention-based Models

Table 2 reports the exponentiated coefficients of the last fully connected layer of the "Attention (concatenated time)" model. They can be interpreted as odds ratios for readmission within 30 days from ICU discharge. Length of stay in the ICU was not associated with readmission risk; however, a longer length of stay before admission to the ICU had a minor protective effect on the odds of experiencing readmission, with an expected OR of 0.994 (95% credible interval of [0.993, 0.996]) between patients hospitalised for one additional day and other patients with the same values for other covariates. Male gender, a higher number of recent admissions, and older age were all associated with higher odds of readmission (OR: 1.114 [1.092, 1.136], 1.187 [1.170, 1.205], and 1.009 [1.009, 1.010], respectively) whereas admission for elective surgery was associated with lower odds of readmission (OR: 0.941 [0.891, 0.993]). Patients with admission location physician referral/normal delivery (OR: 0.882 [0.844, 0.922]) had lower odds of readmission compared with patients admitted through the emergency department. Patients insured through other government programs, with private health insurance, or who self-paid the hospitalization had lower odds of readmission compared with Medicare patients (OR: 0.775 [0.694, 0.865], 0.820 [0.798, 0.843], 0.559 [0.447, 0.700], respectively); however, the odds were similar for Medicare and Medicaid patients (OR: 0.997 [0.992, 1.002]). Marital status was not associated with a difference in odds. Black/African American patients were more likely to experience readmission than white patients (OR: 1.165 [1.118, 1.215]).

Scores for individual diagnosis, procedure, and medication codes associated with an increased risk of readmission within 30 days of discharge from the ICU are reported in Table 3. Patients-at-risk included subjects suffering from infectious complications, e.g. following the insertion of cardiac devices or venous catheters. Similarly, patients diagnosed with specific chronic (uncontrolled type I diabetes, liver disease) or progressive (hepatorenal syndrome) conditions were also at increased risk of early readmission. Another group of interest included patients for whom standard medical care is not possible, e.g. due to contraindications to surgery or medications, such as hydantoin derivatives (a class of anticonvulsants), and requiring desensitization to allergens. The high scores assigned by the model to diagnosed dysphagia and gastrostomy procedures may be associated with critically ill patients. Vital signs associated with increased risk of readmission included a recorded body temperature in the range 33.22-35.93 °C (score: 1.7 [0.2, 3.2]), mean arterial pressure in the range 51-61.32 mmHg (score: 1.2 [0.7, 1.7]), and a respiratory rate in the range 31-44 breaths per minute (score: 1.1 [0.2, 1.9]).

Table 2. The exponentiated coefficients of the last fully connected layer of the "Attention (concatenated time)" model can be interpreted as odds ratios for experiencing an adverse outcome following discharge from the intensive care unit, similarly to traditional logistic regression. Patients with gender: "female", ethnicity: "white", marital status: "married/life partner", insurance: "Medicare", admission location: "emergency room admit" constitute the reference group. Asterisks indicate that the odds ratio's credible interval (CI) does not include one.

| OR [95% CI] | Covariate |
|---|---|
| 1.000 [0.998, 1.002] | ICU Length of Stay (days) |
| 1.114 [1.092, 1.136]* | Gender: Male |
| 1.187 [1.170, 1.205]* | Number of Recent Admissions |
| 1.009 [1.009, 1.010]* | Age (years) |
| 0.994 [0.993, 0.996]* | Pre-ICU Length of Stay (days) |
| 0.941 [0.891, 0.993]* | Elective Surgery |
| 0.998 [0.992, 1.003] | Admission Location: Clinic Referral/Premature Delivery |
| 1.639 [1.146, 2.345]* | Admission Location: Other/Unknown |
| 0.882 [0.844, 0.922]* | Admission Location: Physician Referral/Normal Delivery |
| 1.115 [1.074, 1.157]* | Admission Location: Transfer from Hospital/Extramural |
| 1.001 [0.996, 1.006] | Admission Location: Transfer from Skilled Nursing Facility |
| 0.775 [0.694, 0.865]* | Insurance: Government |
| 0.997 [0.992, 1.002] | Insurance: Medicaid |
| 0.820 [0.798, 0.843]* | Insurance: Private |
| 0.559 [0.447, 0.700]* | Insurance: Self Pay |
| 0.918 [0.845, 0.997]* | Marital Status: Other/Unknown |
| 1.000 [0.995, 1.005] | Marital Status: Single |
| 0.996 [0.991, 1.001] | Marital Status: Widowed/Divorced/Separated |
| 0.772 [0.694, 0.858]* | Ethnicity: Asian |
| 1.165 [1.118, 1.215]* | Ethnicity: Black/African American |
| 1.001 [0.996, 1.006] | Ethnicity: Hispanic/Latino |
| 0.873 [0.832, 0.916]* | Ethnicity: Other/Unknown |
| 1.000 [0.995, 1.004] | Ethnicity: Unable to Obtain |
| 3.780 [3.663, 3.902]* | Score: Diagnoses and Procedures |
| 2.044 [1.979, 2.110]* | Score: Medications and Vital Signs |

Table 3. ICD-9 diagnosis and procedure codes and medications assigned high scores by the "Attention (concatenated time)" model; a high positive score corresponds to increased risk of readmission to the intensive care unit. The final diagnoses/procedures/medications scores for each patient are computed as a weighted average of the scores associated with each individual item. CI: credible interval; ICD: international classification of diseases and related health problems.

| Score [95% CI] | ICD-9 Diagnoses |
|---|---|
| 7.5 [4.8, 10.3] | Infection and inflammatory reaction due to cardiac device, implant, and graft |
| 6.9 [4.7, 9.1] | Other and unspecified infection due to central venous catheter |
| 6.5 [5.3, 7.6] | Need for desensitization to allergens |
| 6.2 [4.5, 7.9] | Hepatorenal syndrome |
| 5.8 [3.9, 7.7] | Diabetes with renal manifestations, type I [juvenile type], uncontrolled |
| 5.4 [3.1, 7.9] | Hydantoin derivatives causing adverse effects in therapeutic use |
| 5.4 [3.4, 7.4] | Encounter for palliative care |
| 5.3 [3.0, 7.7] | Dysphagia, oropharyngeal phase |
| 5.2 [2.7, 8.2] | Spontaneous bacterial peritonitis |
| 5.0 [2.4, 7.4] | Other sequelae of chronic liver disease |
| **Score [95% CI]** | **ICD-9 Procedures** |
| 6.9 [5.0, 9.0] | Other gastrostomy |
| 6.1 [4.8, 7.4] | Therapeutic plasmapheresis |
| 5.6 [2.7, 8.4] | Incision of abdominal wall |
| 4.9 [3.0, 7.0] | Transcatheter embolization for gastric or duodenal bleeding |
| 4.8 [3.0, 6.7] | Transfusion of coagulation factors |
| 4.4 [2.4, 6.4] | Graft of muscle or fascia |
| 4.3 [2.7, 6.0] | Cardiopulmonary resuscitation, not otherwise specified |
| 4.2 [2.7, 5.6] | Endovascular implantation of other graft in abdominal aorta |
| 4.1 [3.0, 5.2] | Reopening of recent thoracotomy site |
| 4.0 [1.6, 6.3] | Other percutaneous procedures on biliary tract |
| **Score [95% CI]** | **Medications** |
| 4.7 [3.3, 6.2] | D5W |
| 4.6 [2.1, 7.2] | Phytonadione |
| 4.2 [2.1, 6.4] | 5% Dextrose |
| 3.8 [2.2, 5.3] | Furosemide |
| 3.4 [1.7, 5.1] | Albuterol 0.083% neb soln |
| 3.3 [1.8, 4.7] | Heparin Sodium |
| 3.2 [1.7, 4.8] | Lorazepam |
| 3.2 [1.3, 5.0] | Hydralazine |
| 3.1 [1.5, 4.8] | 0.9% Sodium Chloride |
| 2.7 [0.7, 4.8] | Acetylcysteine20% |

## Discussion

### Principal Results

This study benchmarked several deep learning architectures for processing time-series sampled at irregular intervals, including a novel application of neural ODEs to model the dynamics in time of medical code embeddings. Models were trained to predict the risk of readmission within 30 days of discharge from the ICU using the MIMIC-III data set. In general, the predictive accuracy of the different deep learning models was similar, but considerably better than logistic regression. It is possible that diagnosis and procedure codes associated with ICU admissions prior to the index admission, as well as medications and vital signs, provide limited additional value when predicting readmission, therefore limiting the impact of different network architectures on predictive accuracy. Neural ODEs applied to code embeddings did generally result in improved performance, suggesting that they may constitute a building block of interest for neural networks processing not only continuous time series, but also timestamped codes.

Models with a recurrent component performed marginally better in terms of accuracy than models based solely on attention mechanisms; these findings are consistent with a previous study which predicted clinical outcomes using the MIMIC-III data [12]. However, by formulating an attention-based model in a Bayesian setting it was possible to evaluate not only predictive performance [12,47] but also to derive credible intervals on network weights and risk predictions; as well as to provide a high degree of interpretability for all model coefficients including those encoding the longitudinal aspect of EMR data. Since training attention-based networks is more efficient than training RNNs, it should be possible to scale the proposed architecture to larger data sets and frequently update network weights as new data is collected within hospitals. Accuracy could be improved further by constructing a Bayesian ensemble of classifiers [37].

## Comparison with Prior Work

The interpretation of the attention-based model supports several previous studies which identified associations between increased risk of readmission and male gender, older age, and admission location [25,48-50]. Whereas previous work found that length of ICU stay was higher among readmitted patients, the present study was inconclusive in this regard [25,49,50]. Similarly to the OASIS severity of illness score, a higher risk of readmission was predicted for patients with a very short hospital length of stay before admission to the ICU and for patients who were not admitted for elective surgery [29].

This study also identified patients suffering from infectious complications [51,52] or chronic conditions, such as diabetes [53] or cirrhosis [25,31], as being at increased risk of readmission. Further, this study emphasizes that patients for whom standard medical care is not possible, e.g. due to contraindications to surgery or medications, may also be at increased risk of readmission, a finding which should be examined further by future studies.

Compared with previous work on predicting general hospital readmission, this study also identified a significant risk for patients belonging to minority groups [54,55] but not for patients living alone [55,56]. The latter finding may reflect the fact that patients are usually discharged from the ICU to a hospital ward, rather than directly to home. Finally, Medicaid and Medicare patients may be at increased risk of ICU readmission compared with privately insured patients [56,57].

## Limitations

The present study has several limitations. Since all dates in the MIMIC-III data set were shifted to protect patient confidentiality, it was not possible to ascertain which patients were admitted after 2001 and had at least 12 months of prior data, possibly leading to some incorrect values for the number of ICU admissions in the year preceding discharge. The use of attention mechanisms with multiple keys/values or multiple attention heads [11] was not assessed; however, these algorithms might be more relevant for networks with a decoder component or multi-class/multi-label classification tasks. The proposed model does not address interactions between static variables or non-linear associations between static variables and predicted risk. It also doesn't account for within-patient clustering of multiple ICU admissions. Further, information from clinical notes [58] was not included and the simplifying assumption was made that various diagnosis- and procedure-related codes were available immediately at the time of discharge. Including prior medical knowledge in the model is currently not possible (however, as expected, normal vital signs were assigned a very low risk score by the attention-based model). A larger prospective study using Australian hospital data will be used to address these limitations.

## Conclusions

In conclusion, this study compared several deep-learning models for predicting the risk of readmission within 30 days of discharge from the ICU based on the full clinical history of a patient. The considered models included a novel application of neural ODEs to model how the predictive relevance of medical codes changes over time. Training an attention-based network in a Bayesian setting allowed insights into intensive-care patients at increased risk of readmission. Overall, attention-based networks may be preferable to recurrent networks if an interpretable model is required, at only marginal cost in predictive accuracy. The development of interpretable machine learning techniques such as proposed here is necessary to allow the integration of predictive models in clinical processes.

## Data Availability

The MIMIC-III data set used in this study is publicly available [32].

Supplementary Information

Supplementary Table S1. Baseline characteristics of the analysed intensive care unit (ICU) stays.

Abbreviations

EMR: electronic medical record

ICU: intensive care unit

ICD-9: international classification of diseases and related health problems version 9

OR: odds ratio

MCE: medical concept embedding

RNN: recurrent neural network

ODE: ordinary differential equation

Contributions

S.B. conceived and designed the analysis. S.B. and J.K. performed the analysis. S.K., M.G., A.R. provided clinical input. O.P.-C. and L.J. provided input on the analytical methods. S.B. wrote the manuscript with the help of all other authors.

Competing Interests

The authors declare no competing interests.

## Figures

Figure 1. Overview of the considered neural network architectures.

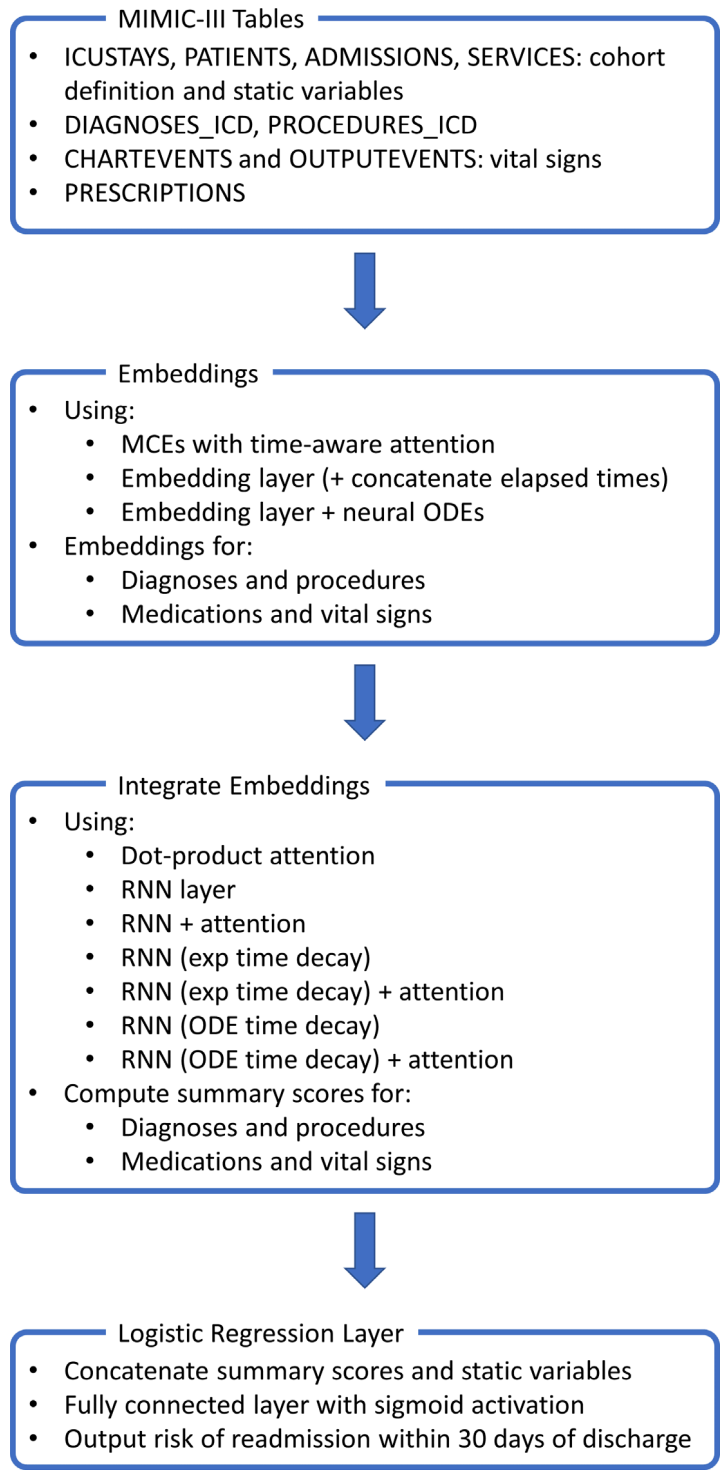

Supplementary Information

Supplementary Table S1. Baseline characteristics of the analysed intensive care unit (ICU) stays.

|  | Readmission Within 30 Days | No Readmission Within 30 Days |
|---|---|---|
|  | N=5,495 | N=39,803 |
| **Num. Recent Admissions** (Mean [Range]) | 0.7 [0-23] | 0.3 [0-12] |
| **ICU Length of Stay** (Days) (Mean [Range]) | 5 [0-117] | 4 [0-173] |
| **Pre-ICU Length of Stay** (Days) (Mean [Range]) | 3 [0-141] | 2 [0-155] |
| **Age** (Years) (Mean [Range]) | 65 [18-100] | 63 [18-103] |
| **Gender** Male | 3136 (57.1 %) | 22536 (56.6 %) |
| **Admission Location** | | |
| Emergency Room Admit | 2537 (46.2 %) | 17251 (43.3 %) |
| Clinic Referral / Premature | 1086 (19.8 %) | 7950 (20.0 %) |
| Transfer from Hospital / Extramural | 1011 (18.4 %) | 6546 (16.4 %) |
| Phys Referral / Normal Delivery | 797 (14.5 %) | 7843 (19.7 %) |
| Transfer from Skilled Nursing Facility | 44 (0.8 %) | 168 (0.4 %) |
| Other / Unknown | 20 (0.4 %) | 45 (0.1 %) |
| **Ethnicity** | | |
| White | 3974 (72.3 %) | 28508 (71.6 %) |
| Black / African American | 679 (12.4 %) | 3806 (9.6 %) |
| Other / Unknown | 489 (8.9 %) | 4523 (11.4 %) |
| Hispanic / Latino | 174 (3.2 %) | 1462 (3.7 %) |
| Asian | 116 (2.1 %) | 920 (2.3 %) |
| Unable to Obtain | 63 (1.1 %) | 584 (1.5 %) |
| **Insurance** | | |
| Medicare | 3328 (60.6 %) | 20850 (52.4 %) |
| Private | 1487 (27.1 %) | 13596 (34.2 %) |
| Medicaid | 550 (10.0 %) | 3676 (9.2 %) |
| Government | 109 (2.0 %) | 1222 (3.1 %) |
| Self-Pay | 21 (0.4 %) | 459 (1.2 %) |
| **Marital Status** | | |
| Married / Life Partner | 2604 (47.4 %) | 19215 (48.3 %) |
| Single | 1511 (27.5 %) | 10471 (26.3 %) |
| Widowed / Divorced / Separated | 1182 (21.5 %) | 8248 (20.7 %) |
| Other / Unknown | 198 (3.6 %) | 1869 (4.7 %) |
| **Elective Surgery Admission** Yes | 513 (9.3 %) | 5729 (14.4 %) |